\documentclass{article}
\usepackage{amsmath,graphicx}
\usepackage{booktabs}
\usepackage{hyperref}					
\hypersetup{colorlinks,
	linkcolor=black,%
	citecolor=black}
\usepackage{ICASSP2021}
\usepackage{multirow}

\title{MixSpeech: Data Augmentation for Low-resource Automatic Speech Recognition}
\name{Linghui Meng\textsuperscript{\rm 1,2}, Jin Xu\textsuperscript{\rm 3}, Xu Tan\textsuperscript{\rm 4}, Jindong Wang\textsuperscript{\rm 4}, Tao Qin\textsuperscript{\rm 4}, Bo Xu\textsuperscript{\rm 1,2}\thanks{This work is supported by the National Key Research and Development Program of China under No.2017YFB1002102}}
\address{\textsuperscript{\rm 1}Institute of Automation, Chinese Academy of Sciences, China\\
\textsuperscript{\rm 2}School of Artificial Intelligence, University of Chinese Academy of Sciences, China\\
\textsuperscript{\rm 3} Institute for Interdisciplinary Information Sciences, Tsinghua University, China \\
\textsuperscript{\rm 4} Microsoft Research Asia\\
\{menglinghui2019,xubo\}@ia.ac.cn, j-xu18@mails.tsinghua.edu.cn, \\ \{xuta,jindong.wang,taoqin\}@microsoft.com}

\begin{document}
%
\maketitle
\begin{abstract}
In this paper, we propose MixSpeech, a simple yet effective data augmentation method based on mixup for automatic speech recognition (ASR). MixSpeech trains an ASR model by taking a weighted combination of two different speech features (e.g., mel-spectrograms or MFCC) as the input, and recognizing both text sequences, where the two recognition losses use the same combination weight. We apply MixSpeech on two popular end-to-end speech recognition models including LAS (Listen, Attend and Spell) and Transformer, and conduct experiments on several low-resource datasets including TIMIT, WSJ, and HKUST. Experimental results show that MixSpeech achieves better accuracy than the baseline models without data augmentation, and outperforms a strong data augmentation method SpecAugment on these recognition tasks. Specifically, MixSpeech outperforms SpecAugment with a relative PER improvement of 10.6$\%$ on TIMIT dataset, and achieves a strong WER of 4.7$\%$ on WSJ dataset.
\end{abstract}
\section{Introduction}
\label{sec:intro}

Automatic speech recognition (ASR) has achieved rapid progress with the development of deep learning. Advanced models such as DNN~\cite{dahl2011context}, CNN~\cite{sainath2013deep}, RNN~\cite{graves2013speech} and end-to-end models~\cite{chan2016listen,mohamed2019transformers} result in better recognition accuracy compared with conventional hybrid models~\cite{trentin2001survey}. However, as a side effect, deep learning-based models require a large amount of labeled training data to combat overfitting and ensure high accuracy, especially for speech recognition tasks with few training data. 

Therefore, a lot of data augmentation methods for ASR~\cite{ko2015audio,shahnawazuddin2016pitch,toth2018perceptually,park2019specaugment,ren2019almost,xu2020lrspeech} were proposed, mainly on augmenting speech data. For example, speed perturbation \cite{ko2015audio}, pitch adjust~\cite{shahnawazuddin2016pitch}, adding noise~\cite{toth2018perceptually} and vocal tract length perturbation increases the quantity of speech data by adjusting the speed or pitch of the audio, or by adding noisy audio on the original clean audio, or by transforming spectrograms. Recently, SpecAugment~\cite{park2019specaugment} was proposed to mask the mel-spectrogram along the time and frequency axes, and achieve good improvements on recognition accuracy.  Furthermore, \cite{wang2019semantic} masks the speech sequence in the time domain according to the alignment with text to explore the semantical relationship. As can be seen, most previous methods focus on augmenting the speech input while not changing the corresponding label (text), which needs careful tuning on the augmentation policy. For example, SpecAugment needs a lot of hyper-parameters (the time warp parameter $W$, the time and frequency mask parameters $T$ and $F$, a time mask upper bound $p$, and the number of time and frequency mask $m_F$ and $m_T$) to determine how to perform speech augmentation, where improper parameters may cause much information loss and thus cannot generate text correctly, or may have small changes on the speech and thus have no augmentation effect.

Recently, mixup technique~\cite{zhang2017mixup} is proposed to improve the generalization limitation of empirical risk minimization (ERM)~\cite{vapnik1999overview}. Vanilla mixup randomly chooses a weight from a distribution and combines two samples and corresponding labels with the same weight. Recent works apply mixup on different tasks, including image classification~\cite{zhang2018deep}, and sequence classification~\cite{guo2019augmenting}. Different from classification task with one-hot labels where mixup can be easily incorporated, conditional sequence generation tasks such as automatic speech recognition, image captioning, and handwritten text recognition cannot directly apply mixup on the target sequence due to different lengths. 
\cite{medennikov2018investigation} directly apply mixup to train a neural acoustic model like LF-MMI \cite{povey2016purely}, which simply integrates inputs and targets frame-by-frame with shared mixture weight because the input feature and label are aligned frame-wisely.

In this paper, we propose MixSpeech, a simple yet effective data augmentation method for automatic speech recognition. MixSpeech trains an ASR model by taking a weighted combination of two different speech sequences (e.g., mel-spectrograms or MFCC) as input, and recognizing both text sequences. Different from mixup~\cite{zhang2017mixup} that uses the weighted combination of two labels as the new label for the mixed input, MixSpeech uses each label to calculate the recognition loss and combines the two losses using the same weight as in the speech input. MixSpeech is much simple with only a single hyper-parameter (the combination weight $\lambda$), unlike the complicated hyper-parameters used in SpecAugment. Meanwhile, MixSpeech augments the speech input by introducing another speech, which acts like a contrastive signal to force the ASR model to better recognize the correct text of the corresponding speech instead of misled by the other speech signal. Therefore, MixSpeech is more effective than previous augmentation methods that only change the speech input with masking, wrapping, pitch, and duration adjusting, without introducing contrastive signals. 

\section{Method}
In this section, we first briefly recap the concept of mixup~\cite{zhang2017mixup}, which is an efficient augmentation approach for the single-label task. In order to handle sequence generation tasks (e.g., ASR), then we introduce MixSpeech. At the last, we describe two popular models on which MixSpeech is implemented to verify the effectiveness of MixSpeech.  

\subsection{Mixup Recap}
Mixup~\cite{zhang2017mixup} is an effective data augmentation method for supervised learning tasks. It trains a model on a convex combination of pairs of inputs and their targets to make the model more robust to adversarial samples. The mixup procedure can be demonstrated as:
\begin{equation}
\begin{aligned}
    \label{eq:data_mix}
    X_{mix} &= \lambda X_i + (1-\lambda) X_j,\\
    Y_{mix} &= \lambda Y_i + (1-\lambda) Y_j,
\end{aligned}
\end{equation}
where $X_i$ and $Y_i$ are the $i$-th input and target of the data sample, $X_{mix}$ and $Y_{mix}$ represent the mixup data by combining a pair of data samples ($i$ and $j$), and $\lambda \sim Beta(\alpha, \alpha)$ with $\alpha\in(0, \infty)$, is the combination weight.

\subsection{MixSpeech}

Mixup can be easily applied to classification tasks where the target is a single label. However, it is difficult to apply into sequence generation tasks such as ASR due to the following reasons: 1) Two text sequences ($Y_i$, $Y_j$) may have different lengths and thus cannot be directly mixed up as in Eq.~\ref{eq:data_mix}; 2) Text sequences are discrete and cannot be directly added; 3) If adding two text sequences in the embedding space, the model will learn to predict a mixture embedding of two text sequences, which will confuse the ASR model and may hurt the performance since its final goal is to recognize a single text from the speech input. 

\begin{figure}[t]
  \centering
  \includegraphics[scale=0.3]{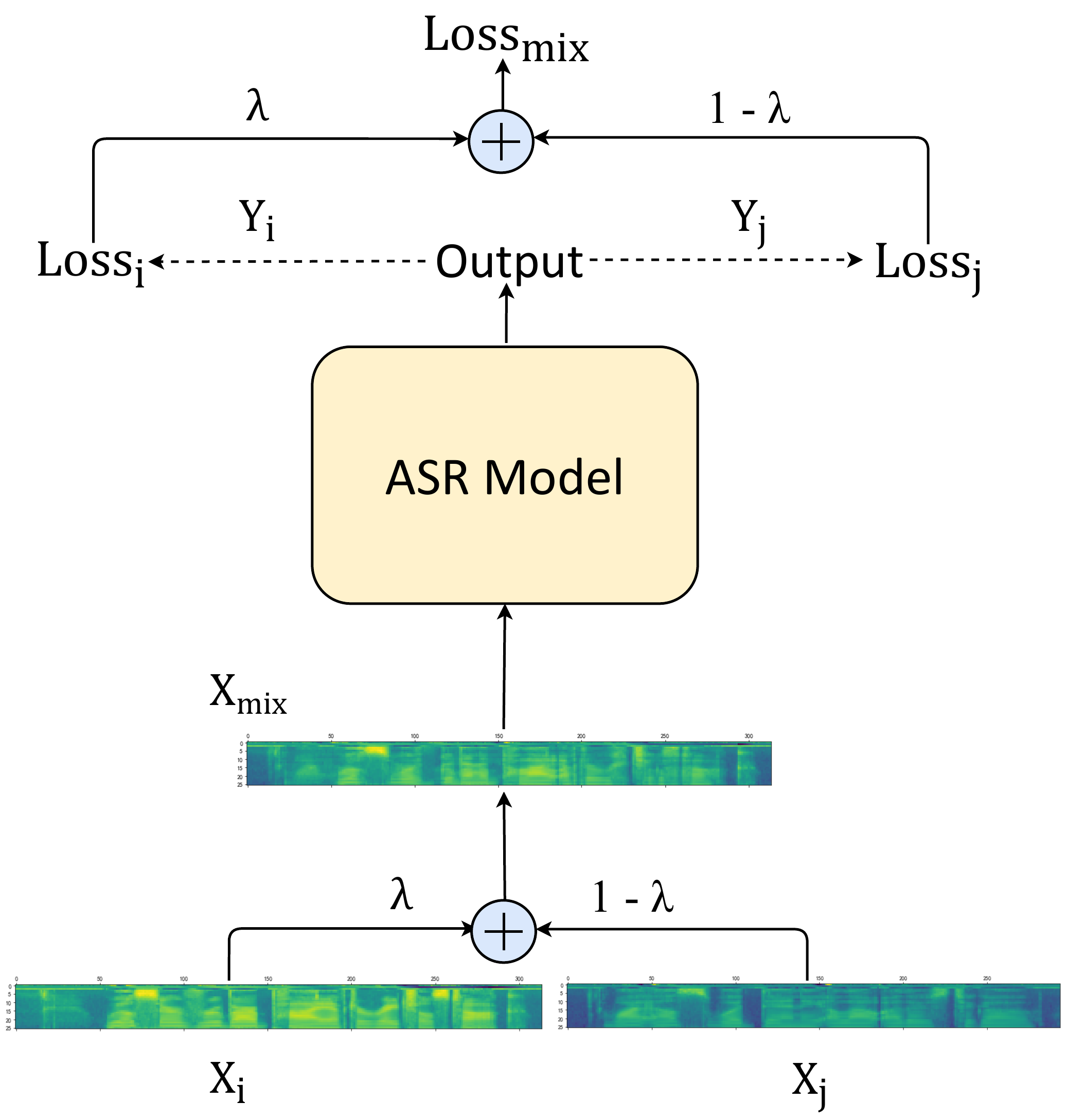}
  \caption{The pipeline of MixSpeech. $X_{mix}$ is the mixture of two mel-spectrograms, $X_i$ and $X_j$. $Loss_i$ is calculated from the output of $X_{mix}$ and the corresponding label $Y_i$. $Loss_{mix}$ is the weighted combination of two losses $Loss_i$ and $Loss_j$.}
  \label{fig:model_structure}
\end{figure}

To ensure effective mixup for the sequential data (speech and text) in ASR, we propose MixSpeech, which mixes two speech sequences in the input and mixes two loss functions regarding the text output, as shown in Figure \ref{fig:model_structure}. The formulation of MixSpeech is as follows: 

\begin{equation}
    \begin{aligned}\label{eq:mixspeech}
    & X_{mix} = \lambda X_i + (1-\lambda) X_j, \\
    &\mathcal{L}_{i} = \mathcal{L}(X_{mix},Y_i) \quad \mathcal{L}_{j} = \mathcal{L}(X_{mix},Y_j), \\
    &\mathcal{L}_{mix}=\lambda\mathcal{L}_{i} + (1-\lambda)\mathcal{L}_{j},
    \end{aligned}
\end{equation}
where $X_i$ and $Y_i$ are the input speech sequence and target text sequence of $i$-th sample, $X_{mix}$ is the mixup speech sequence by adding the two speech sequences frame-wisely with weight $\lambda$. $\mathcal{L}(\cdot, \cdot)$ calculates the ASR loss (which also includes the recognition process), and $\mathcal{L}_{mix}$ combines the two losses with the same weight $\lambda$ as in the speech input during the training phase. Following the original mixup, we choose $\lambda \sim Beta(\alpha, \alpha)$ where $\alpha\in(0, \infty)$.

\subsection{Model Structure}

LAS~\cite{chan2016listen} and Transformer~\cite{vaswani2017attention} architectures have been widely used and achieved great performance on ASR tasks. In this paper, we implement MixSpeech on these two popular models to demonstrate its effectiveness. The two models both leverage the joint CTC-attention~\cite{kim2017joint} structure to train the model, which consists of two loss functions: 1) CTC (Connectionist Temporal Classification) \cite{graves2006connectionist} loss on the encoder output and 2) Cross-Entropy (CE) loss at the output of decoder at each timestep. To train the CTC part, a sequence of labels, denoted as $l$, are used to compute loss with an effective algorithm such as the Baum-Welch algorithm~\cite{baum1970maximization}. The CTC loss can be written as: 

\begin{equation}
    \mathcal{L}_{CTC}(\hat{y},l) = -\log\left(\sum_{\pi\in\beta^{-1}(l)}\prod_{t=1}^T y_{\pi_t}^t\right),
\end{equation}
where $\beta$ is the function that removes repeated labels and blank labels, $\pi_t$ represents intermediate label sequence containing blank label and $l$ is the target sequence without blank. The Cross-Entropy loss can be written as:

\begin{equation}
\begin{aligned}
    \mathcal{L}_{CE} &=-\sum_{u}\log(P(y_u|x, y_{1:u-1})),\\
\end{aligned} 
\end{equation}
where $y_u$ is the $u$-th target token. In LAS~\cite{chan2016listen}, the encoder and decoder are the stacked BiLSTM and LSTM respectively. In Transformer based structure~\cite{vaswani2017attention}, they are stacked of multi-head attention and feed-forward network.

During training, following~\cite{kim2017joint}, we leverage the multi-task learning by combining CTC and Cross-Entropy loss as follows: 

\begin{equation}
\begin{aligned}
    \label{eq:multi-task-learning}
    \mathcal{L}_{MTL} = \beta \mathcal {L}_{CTC} + &(1-\beta) \mathcal{L}_{CE},
\end{aligned}
\end{equation}
where $\beta \in [0,1]$ is a tunable hyper-parameter. By combining with MixSpeech in Eq.~\ref{eq:mixspeech}, the final training objective can be written as: 

\begin{equation}
\label{eq:mixup_para}
    \mathcal{L}_{mix} = \lambda\mathcal{L}_{MTL}(X_{mix},Y_i) + (1-\lambda)\mathcal{L}_{MTL}(X_{mix},Y_j).
\end{equation}

\section{Experimental results}
\subsection{Datasets}

We evaluate our proposed MixSpeech on several datasets including TIMIT~\cite{garofolo1993darpa},  WSJ~\cite{paul1992design} and the Mandarin ASR corpus HKUST~\cite{liu2006hkust}. For TIMIT dataset, we use the standard train/test split, where we randomly sample 10$\%$ of the training set for validation and regard the left as the training set. WSJ~\cite{paul1992design} is a corpus of read news articles with approximately 81 hours of clean data. Following the standard process, we set the Dev93 for development and Eval92 for evaluation. For HKUST dataset, the development set contains 4,000 utterances (5 hours) extracted from the original training set with 197,387 utterances (173 hours) and the rest are used for training, besides 5,413 utterances (5 hours) for evaluation. The speech sentences of these corpora with 16k sampling rate extract the fbank features as the model input. For TIMIT dataset, fbank features with 23 dimensions are extracted at each frame, while for the other two datasets 80 dimensions fbank features are extracted at each frame following the common practice.

\subsection{Model Settings} \label{sec:settings}

We implement MixSpeech based on the ESPnet codebase\footnote{https://github.com/espnet/espnet}. MixSpeech is designed for extracting and splitting targets from mixed inputs to train a more generalizable model, which is a more difficult task than training without MixSpeech. Thus we give more training time to the model enhanced with MixSpeech\footnote{More training time for the baseline model cannot get better results according to our preliminary experiments.}. 
The baselines of LAS~\cite{chan2016listen} and Transformer~\cite{vaswani2017attention} are implemented as the following settings. For the LAS (Listen Attend and Spell), the encoder has 4-layer Bi-LSTM with width 320 and the decoder has a single layer LSTM with width 300. For the Transformer based architecture, the encoder has 12 layers and the decoder has 6 layers with width 256, where each layer is composed of multi-head attention and fully connected layer. CTC and decoder cross-entropy loss are used for jointly decoding using beam search with beam size 20. The default $\alpha$ for Beta distribution is set to 0.5 following~\cite{zhang2017mixup}. The multi-task learning parameter $\beta$ is set to 0.3. In the practice, we randomly select part of paired data within one batch to train the model with MixSpeech and other data within one batch to train the model as usual. The proportion of one batch data to train using MixSpeech is denoted as $\tau$, which is set to 15\% as default and achieve good performance according to our experiments in Table~\ref{tab:mixup_prob_comparison}.

\subsection{Results of MixSpeech} \label{exp:mixspeech_main}
We compare MixSpeech with two settings: 1) models trained without MixSpeech (Baseline) and 2) models trained with SpecAugment~\cite{park2019specaugment} (SpecAugment), an effective data augmentation technique. Table \ref{tab:preliminary_comparison} shows the performance of MixSpeech on TIMIT dataset with the LAS model, comparing with Baseline and SpecAugment. Furthermore, Table \ref{tab:preliminary_comparison_trans} shows the comparison of our method with Baseline and SpecAugment on three different low resource datasets by different metrics with Transformer. We can find that MixSpeech improves the baseline model on TIMIT dataset relatively by 10.6\% on LAS and 6.9\% on Transformer. As shown in the Table \ref{tab:preliminary_comparison_trans}, MixSpeech achieves 4.7\% WER on WSJ, outperforming the baseline model alone and the baseline model with SpecAugment. MixSpeech also achieves better performance on HKUST~\cite{liu2006hkust}, which is a large Mandarin corpus. The results demonstrate that MixSpeech can consistently improve the recognition accuracy across datasets with different languages and data sizes.

\begin{table}[th]
  \caption{The phone error rate (PER) results of LAS (Listen, Attend and Spell), and LAS enhanced with SpecAugment or MixSpeech on TIMIT dataset.}
  \label{tab:preliminary_comparison}
  \centering
\vspace{3mm}
\begin{tabular}{@{}llllc@{}}
\toprule
\textbf{Method} &  &  &  & \textbf{PER}    \\ \midrule
Baseline        &  &  &  & 21.8\%          \\ 
+SpecAugment     &  &  &  & 20.5\%          \\ \midrule
+MixSpeech       &  &  &  & \textbf{19.5\%} \\ \bottomrule
\end{tabular}
\end{table}

\begin{table}[th]
\caption{The results of Transformer, and Transformer enhanced with SpecAugment or MixSpeech on TIMIT, WSJ, and HKUST dataset. The metrics of the three datasets are phone error rate (PER) and word error rate (WER) respectively.}
  \label{tab:preliminary_comparison_trans}
  \centering
\vspace{3mm}
\begin{tabular}{@{}lccc@{}}
\toprule
$\textbf{Method}$ & $\textbf{TIMIT}_{PER}$ & $\textbf{WSJ}_{WER}$ & $\textbf{HKUST}_{WER}$ \\ \midrule
Baseline & 23.1\% & 5.9\% & 23.9\% \\
+SpecAugment & 22.5\% & 5.4\% & 23.5\% \\ \midrule
+MixSpeech & \textbf{21.8\%} & \textbf{4.7\%} & \textbf{22.9\%} \\ \bottomrule
\end{tabular}
\end{table}

\subsection{Method Analysis}

Both LAS and Transformer based models leverage multi-task learning to boost the performance and $\beta$ in Equation~\ref{eq:multi-task-learning} is a hyper-parameter to adjust the weight of them. When $\beta$ is set to 0 or 1, there is only Cross-Entropy loss or CTC loss. We vary $\beta$ and get the results of the baseline as shown in Table~\ref{tab:mtl_comparison}. The results show that the model cannot perform well with only one target and requires alignment information guided by CTC loss to help the attention decoder.  In Table~\ref{tab:mtl_comparison},  $\beta=0.3$ gets the lowest PER score, and thus is set as default in the baselines and models enhanced with MixSpeech.

\begin{table}[th]
  \caption{The phone error rate (PER) results by varying the multi-task learning parameters $\beta$ on TIMIT dataset.}
  \label{tab:mtl_comparison}
  \centering
\vspace{3mm}
\begin{tabular}{@{}ccccc@{}}
\toprule
$\beta$ & 0      & 0.3             & 0.5    & 0.7    \\ \midrule
\textbf{PER}      & 26.6\% & \textbf{23.0\%} & 23.1\% & 23.7\% \\ \bottomrule
\end{tabular}
\vspace{1mm}
\end{table}

As described in Sec.~\ref{sec:settings}, we randomly select part of paired data within one batch (the proportion is denoted as $\tau$) to train the model with MixSpeech and other data within one batch to train the model as usual. We further study whether the proportion $\tau$ of a batch data has much influence on the results and the results are shown in Table~\ref{tab:mixup_prob_comparison}. we can see that the model enhanced with MixSpeech can consistently outperform the baseline ($\tau=0$) by varying $\tau$, which shows that MixSpeech is not sensitive to $\tau$.

\begin{table}[th]
  \caption{The phone error rate (PER) results by varying proportion $\tau$ of a batch data to conduct MixSpeech on TIMIT dataset.}
  \label{tab:mixup_prob_comparison}
  \centering
\vspace{3mm}
\begin{tabular}{@{}ccccc@{}}
\toprule
$\tau$ & 0\%    & 15\%            & 20\%   & 30\%   \\ \midrule
\textbf{PER}        & 23.1\% & \textbf{21.8\%} & 21.9\% & 22.0\% \\ \bottomrule
\end{tabular}
\end{table}

Rather than mixup on two inputs, we can extend mixup to more inputs. Here we randomly select three inputs, named Tri-mixup, and set the mixup weight $\lambda$ to $1/3$\footnote{\cite{zhang2017mixup} points out that complex prior distribution like the Dirichlet distribution to generate $\lambda$ fails to provide further gain but increase the computation cost. Thus we choose $\lambda=$1/3.}. Similar to MixSpeech, the proportion for mixup $\tau$ is also set to 15\%. Table \ref{tab:tri_mixup} shows that the performance of Tri-mixup drops dramatically, comparing with the baseline and MixSpeech. Tri-mixup may introduce too much information beyond the original ones, which hinders the model from learning useful patterns from the data and leads to the performance drop.

\begin{table}[th]
  \caption{The phone error rate (PER) results of different data augmentation methods including Tri-mixup, Noise Regularization and MixSpeech on TIMIT dataset. The Baseline refers to the Transformer model without data augmentation. Tri-mixup means that we select three inputs and conduct mixup with weight $\lambda=$1/3. The noise regularization is to add Gaussian noise to make the model more robust to noise.}
  \label{tab:tri_mixup}
  \centering
\vspace{3mm}
\begin{tabular}{@{}ccccc@{}}
\toprule
             & \multirow{2}{*}{Baseline} & \multirow{2}{*}{Tri-mixup} & Noise & \multirow{2}{*}{MixSpeech} \\
             &  &  & Regularization &  \\ \midrule
\textbf{PER} & 23.1\%            & 35.8\%             & 22.0\%             & \textbf{21.8\%}    \\ \bottomrule
\end{tabular}
\end{table}

Noise regularization \cite{seltzer2013investigation, yu2008minimum, maas2012recurrent}, which incorporates noise on the inputs, is a common data augmentation method to make the model more robust to noise and more generalizable to unseen data. To compare it with MixSpeech, we add Gaussian noise with SNR 5dB on the training set and report the results in Table~\ref{tab:tri_mixup}. We can see that MixSpeech outperforms the noise regularization. 

\section{Conclusion and future work}

In this paper, we proposed MixSpeech, a new data augmentation method to apply the mixup technique on ASR tasks for low-resource scenarios. Experimental results show that our method improves the recognition accuracy compared with the baseline model and previous data augmentation method SpecAgument, and achieves competitive WER performance on WSJ dataset. In the future, we will consider another mixup strategy by concatenating different segments of speech and also their corresponding text sequences, which can create more data samples with novel distributions.

\ninept
\bibliographystyle{IEEEtran}

\bibliography{mybib}

\end{document}